\documentclass[a4paper,10pt]{article}

\usepackage{fullpage}
\usepackage{authblk}
\usepackage[numbers, sort&compress]{natbib}
\usepackage{graphicx}
\usepackage{subcaption}
\usepackage{enumitem}
\usepackage{xspace}
\usepackage{hyperref}
\usepackage{url}
\usepackage{amsthm}
\usepackage{nicefrac}
\usepackage{booktabs}
\usepackage{multirow}

\usepackage{amsmath,amsfonts,bm,amssymb}

\def\eqref#1{equation~\ref{#1}}

\def\1{\bm{1}}

\def\mbx{{\mathbf{x}}}

\def\mbX{{\mathbf{X}}}

\def\mcX{{\mathcal{X}}}
\def\mcY{{\mathcal{Y}}}
\def\mcH{{\mathcal{H}}}
\def\mcG{{\mathcal{G}}}
\def\mcF{{\mathcal{F}}}

\def\mcE{{\mathcal{E}}}

\def\mcS{{\mathcal{S}}}

\DeclareMathAlphabet{\mathsfit}{\encodingdefault}{\sfdefault}{m}{sl}
\SetMathAlphabet{\mathsfit}{bold}{\encodingdefault}{\sfdefault}{bx}{n}

\def\gQ{{\mathcal{Q}}}

\DeclareMathOperator*{\argmax}{arg\,max}
\DeclareMathOperator*{\argmin}{arg\,min}

\newcommand{\tf}{\tilde{f}}
\newcommand{\tg}{\tilde{g}}
\newcommand{\SL}{\ell^{surr}}
\newcommand{\ts}{\tilde{s}}
\newcommand{\indicator}[1]{\mathbb{I}_{\{#1\}}}
\newcommand{\commonfrac}{\frac{1}{2}{\ell}^{surr}(\nicefrac{2}{3})}
\newcommand{\smfy}{\sigma^{\tilde{f}}_y(\mbx)}
\newcommand{\smgy}{\sigma^{\tilde{g}}_y(\mbx,h)}

\newcommand{\Joint}{\textsc{LtA-Joint}\xspace}
\newcommand{\Seq}{\textsc{LtA-Seq}\xspace}

\newcommand{\LZeroOnearg}{L_{0-1}^{ask}\left( f(\mbx),g(\mbx,h),s(\mbx), y \right)}
\newcommand{\LAskarg}{L^{ask}\left( f(\mbx),g(\mbx,h),s(\mbx), y \right)}
\newcommand{\LJAskarg}{\tilde{L}^{ask}\left( \tilde{f}(\mbx),\tilde{g}(\mbx,h),\tilde{s}(\mbx), y \right)}

\newcommand{\chx}{\texttt{X-Rays}\xspace}
\newcommand{\synth}{\texttt{Synth}\xspace}

\usepackage[dvipsnames]{xcolor}
\definecolor{yellowburn}{HTML}{FFF2CC}

\newtheorem{theorem}{Theorem}

\title{To Ask or Not to Ask: Learning to Require Human Feedback}

\author[1]{Andrea Pugnana\thanks{These authors contributed equally to the present work.}} 
\author[2]{Giovanni De Toni$^*$} 
\author[1,3]{Cesare Barbera$^*$} 
\author[4]{Roberto Pellungrini} 
\author[2]{Bruno Lepri} 
\author[1]{Andrea Passerini} 

\affil[1]{DISI, University of Trento, Trento, Italy \texttt{\{andrea.pugnana,cesare.babera,andrea.passerini\}@unitn.it}}
\affil[2]{Fondazione Bruno Kessler, Trento, Italy \texttt{\{gdetoni,lepri\}@fbk.eu}}
\affil[3]{Univeristy of Pisa, Pisa, Italy}
\affil[4]{Scuola Normale Superiore, Pisa, Italy \texttt{roberto.pellungrini@sns.it}}

\date{\vspace{-10mm}}

\begin{document}

\maketitle

\begin{abstract}
Developing decision-support systems that complement human performance in classification tasks remains an open challenge. A popular approach, Learning to Defer (LtD), allows a Machine Learning (ML) model to pass difficult cases to a human expert.
However, LtD treats humans and ML models as mutually exclusive decision-makers, restricting the expert contribution to mere predictions.
To address this limitation, we propose Learning to Ask (LtA), a new framework that handles both when and how to incorporate expert input in an ML model. LtA is based on a two-part architecture: a standard ML model and an enriched model trained with additional expert human feedback, with a formally optimal strategy for selecting when to query the enriched model.
We provide two practical implementations of LtA: a sequential approach, which trains the models in stages, and a joint approach, which optimises them simultaneously.
For the latter, we design surrogate losses with realisable-consistency guarantees.
Our experiments with synthetic and real expert data demonstrate that LtA provides a more flexible and powerful foundation for effective human–AI collaboration.
\end{abstract}

\section{Introduction}
\label{sec:introduction}
Advances in Machine Learning (ML) over the past years have led to drastic improvements across a wide range of tasks, enabling models to surpass human abilities and understanding in specific domains \citep{Esteva, He_2015_ICCV, Taigman_2014_CVPR, Zhang, Sinclair, Matek, rajpurkar2017chexnet}.
However, in many real-world applications, full automation is not desirable due to issues such as systemic bias~\citep{mehrabi2021surveybiasandfairness}, lack of accountability~\citep{kaminski2018binary}, and limited generalization~\citep{liu2021towards}.
As a result, there is a growing interest in developing human-AI collaboration frameworks that incorporate human expert feedback, 
and achieve complementary performance~\citep{Bansal}. 
Consider the case of emergency room triage, where healthcare providers assess and prioritise patients for immediate care.
While data-driven decision support systems have demonstrated impressive performance~\citep{currie2017diagnosing,mullainathan2022diagnosing}, they might have access to less information about the patient than the decision maker.  
Indeed, in certain tasks, expert physicians have been shown to leverage information (\textit{e.g.,} the patient's oral history) not available to a decision support system to make decisions \citep{DBLP:conf/nips/AlurLLRSS23}\footnote{
For instance, it has been shown how physicians may consider additional factors (\textit{e.g.,} anticoagulant use) when diagnosing acute gastrointestinal bleeding (AGIB), which are absent from the standard Glasgow-Blatchford Score \citep{laine2021acg}, a nine-variable metric used to assess AGIB risk. Thus, influencing admit/discharge decisions beyond what any algorithm trained on those variables can capture \citep{DBLP:conf/nips/AlurLLRSS23}.
}. 
Incorporating expert feedback can enhance the model's ability to make accurate diagnoses, \textit{e.g.,} in instances where clinical judgment is essential.

\begin{figure*}[t]
    \centering
    \begin{subfigure}[t]{0.34\textwidth}
        \centering
         \includegraphics[width=\linewidth,height=40mm, keepaspectratio]{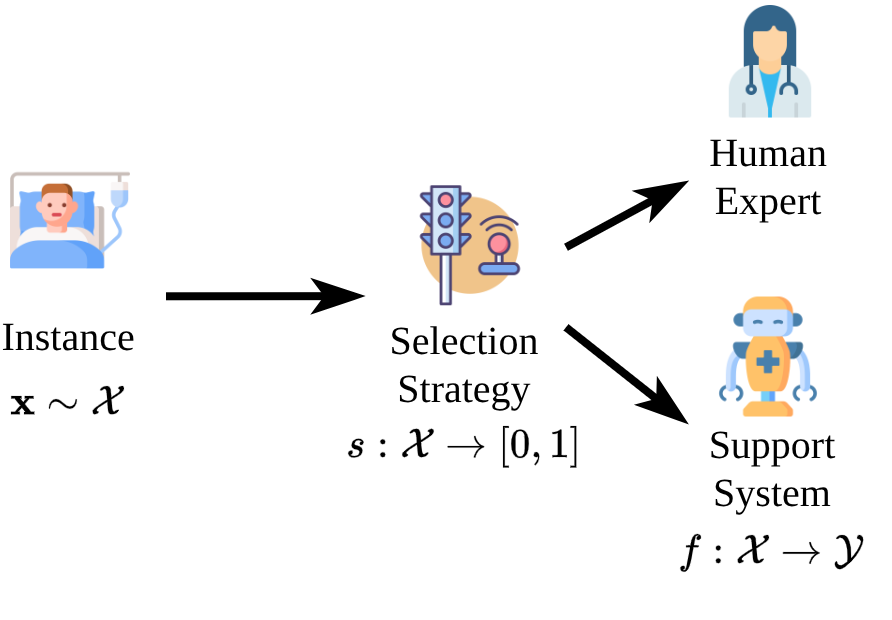}
         \caption{Learning to Defer (LtD)}
    \end{subfigure}
    \hfill
    \begin{subfigure}[t]{0.61\textwidth}
        \centering
         \includegraphics[width=\linewidth,height=40mm, keepaspectratio]{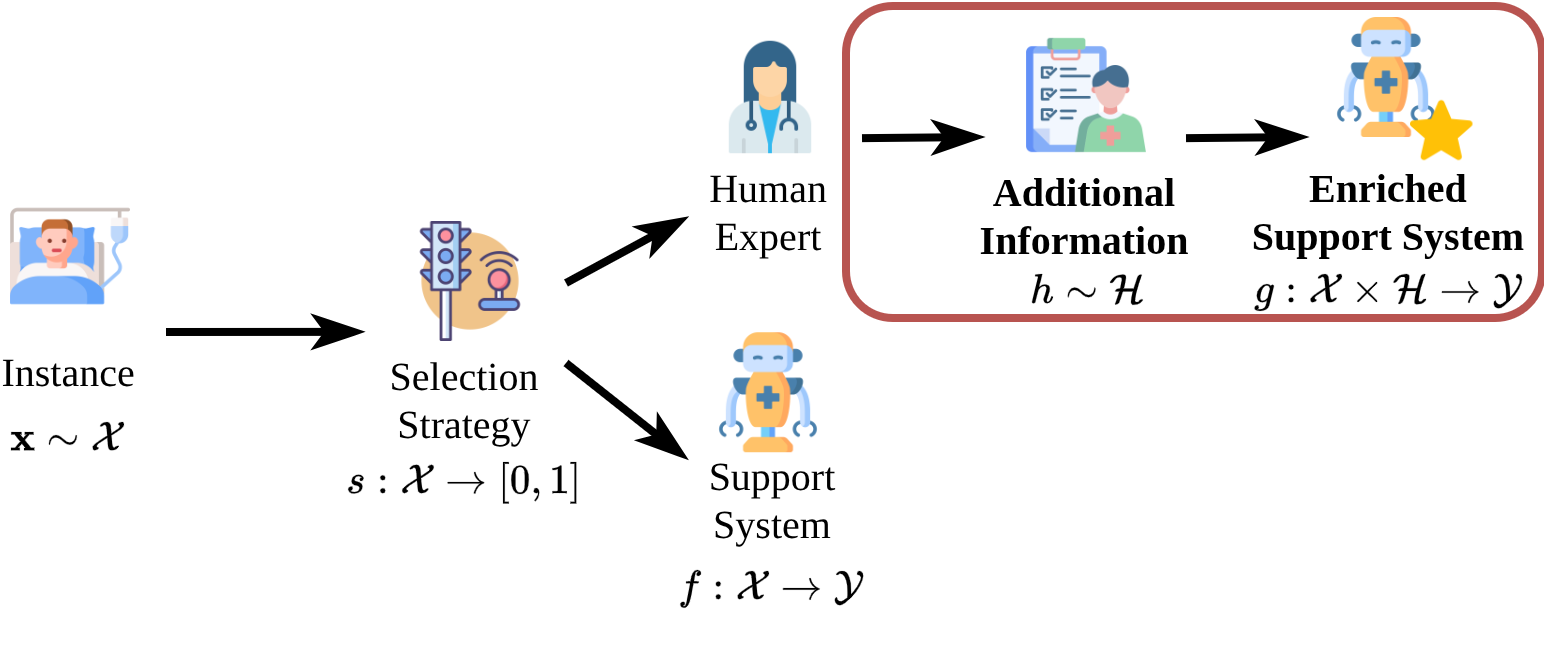}
         \caption{Learning to Ask (LtA)}
    \end{subfigure}
    
    \caption{ \textbf{From Learning to Defer (LtD) to Learning to Ask (LtA).}
    (Left) In classical LtD tasks, given an instance $\mbx \in \mcX$, we employ a selection strategy $s(\mbx)$ to \textit{defer} the final prediction to either a human expert or an ML predictor.
    However, both the expert and the predictor might have access to mutually exclusive informative features (\textit{e.g.,} complex medical signals and the oral medical history of the patient).  
    (Right) Our proposed LtA framework (Section~\ref{sec:LtAformal}) interrogates instead about \textit{when} to request human input, and \textit{how} to incorporate such complementary information $h \in \mcH$ within an enriched predictor $g: \mcX \times \mcH \rightarrow \mcY$.
    }
    \label{fig:lta-framework}
\end{figure*}

The concept of complementary performance has become an important focus in Human-Computer Interaction (HCI)~\citep{lai2021towardshumanai} and, more recently, Hybrid Decision Making (HDM), motivated by the observation that fundamental differences between human reasoning and ML systems allow each to excel on distinct subsets of a task.
However, determining when and how humans and ML systems effectively complement each other remains an open challenge~\citep{steyvers2024three}.
This is particularly critical in high-stakes settings, where systems must be designed to reliably assess whether to rely on the human expert or the machine prediction to maximise overall performance~\citep{vaccaro2024combinations}.
Among various decision-making frameworks, Learning to Defer (LtD)~\citep{DBLP:conf/nips/MadrasPZ18} has recently attracted significant attention.
LtD extends standard supervised learning by introducing a \textit{deferring mechanism}, allowing an ML model to either predict or delegate the decision to a human expert~\citep{DBLP:conf/aaai/RuggieriP25}.
This framework naturally enables more adaptive and collaborative systems, especially in scenarios where human expertise complements algorithmic predictions.

If our goal is complementarity, Learning to Defer (LtD) approaches still have two major shortcomings.
First, LtD methods require human experts to provide task-specific predictions.
However, such a requirement overlooks other forms of expert feedback, which can be leveraged to improve the final task prediction, \textit{e.g.}, concept annotations or partial feature descriptors.
Second, LtD frameworks treat the expert and the machine as \textit{independent} decision-makers, failing to exploit existing synergies between them, and often resulting in systematic underfitting \citep{DBLP:conf/aistats/LiuCZF024}, thereby discarding valuable information.

In this paper, we propose a novel framework, \textbf{Learning to Ask (LtA)}, which addresses key limitations of existing approaches while preserving the flexibility of the LtD paradigm.
Instead of optimising when to delegate the entire decision to either the machine or the human expert, LtA focuses on when to \textit{ask for} expert feedback during the prediction process.
This is achieved through a two-component architecture consisting of a standard classifier and an enriched model that integrates expert input when needed. See Figure~\ref{fig:lta-framework} for a visual representation.
This design not only captures the potential \textit{complementarity} between the two actors but also enables synergistic collaborations, allowing both the machine and the expert to contribute simultaneously.

\paragraph{Our contributions.} First, we show that there exist data distributions (potentially many) for which Learning to Defer (LtD) is provably suboptimal \textit{e.g.,} deferring to either the expert or the machine does not improve the expected accuracy (Section \ref{sec:suboptimal}).
Then, we introduce the formal framework for training ML models that can determine when to ask for expert feedback (Section \ref{sec:LtAformal}), under general additional inputs, and we derive the theoretically optimal \textit{selection strategy} for Learning to Ask (LtA) (Theorem~\ref{thm:optimal_defer}).
Furthermore, we propose two different strategies to learn LtA methods: $(i)$ a sequential model, which first trains the enriched model and then learns the standard model and the policy using well-known LtD losses; $(ii)$ a joint model, which exploits novel realizable-consistent surrogate loss functions that enable joint optimization under LtA (Theorem~\ref{thm:real_consistency}).
Lastly, we empirically validate our proposed strategy on synthetic and real-world classification tasks (Section~\ref{sec:experiments}), showing that LtA outperforms LtD when richer feedback is provided.
We have released an open-source version of the source code used for the experiments and the raw results at the following link \url{https://github.com/andrepugni/LearningToAsk}.

\section{Background}
\label{sec:background}
Let $\mcX$ be a feature space, and let $\mcY$ be the target space.
In LtD, the goal is to learn a model $f:\mcX \to \mcY$ and a selector $s:\mcX\to \{0,1\}$, that can predict the target variable and when to defer to another (human) expert, respectively.
We define the hypothesis space of the predictors and selection strategies as $\mcF\times\mcS$. Further, in our paper, we denote the human expert as another \textit{fixed} predictor $f'\in \mcF'$. 
Given an unknown probability distribution $P'$ over $\mcX\times\mcY\times\mcF'$, the goal of LtD is to minimize the following objective\footnote{To reduce notational burden, we denote with $\mathbb{E}_{v}$ the expectation computed with respect to a random variable $V$ drawn from some distribution $P_V$, \textit{e.g.,} $\mathbb{E}_{(\mbx,y,f')}[\cdot] = \mathbb{E}_{(\mbx,y,f')\sim P'}[\cdot]$.}:
\begin{equation}
    \min_{(f,s) \in \mcF\times\mcS}\mathbb{E}_{(\mbx,y,f')}[L_{0-1}^{def}\left(f(\mbx),s(\mbx), f', y\right)]
    \label{eq:LDEF}
\end{equation}
where $L_{0-1}^{def}\left(f(\mbx),s(\mbx), f', y\right) = \indicator{f(\mbx)\neq y}\indicator{s(\mbx)=0} + (\alpha\indicator{f'\neq y} +\delta)\indicator{s(\mbx) =1}$ is the \textit{deferral 0-1 loss function}, with a normalizing factor $\alpha\in[0,1]$ and defer cost $\delta\in[0,1]$ modeling the cost of querying a human expert.
Unless mentioned, we assume $\alpha=1$ and $\delta=0$, as done in most LtD literature~\citep{DBLP:conf/icml/MozannarS20,DBLP:conf/aistats/MozannarLWSDS23,DBLP:conf/nips/MaoM024b}.
The loss function in Eqn.~(\ref{eq:LDEF}) is generally intractable and hard to directly optimise, hence several works have considered \textit{surrogate losses}.
Formally, let us denote by $\mcE_{L}(q)$ the generalization error of an hypothesis $q\in\gQ$ with respect to loss $L$,  \textit{i.e.,} $\mcE_{L}(q) = \mathbb{E}_{\mbx, y, f'}\left[ L(q(\mbx), y, f')\right]$, and by $\mcE_{L}^*$ the best-in-class generalization error,  \textit{i.e.,} $\mcE_{L}^* = \inf_{q\in \gQ}\mathbb{E}_{\mbx, y, f'}\left[ L(q(\mbx), y, f')\right]$.
In the context of LtD, the hypothesis space $\gQ$ consists of all pairs of predictors $f$ and selection strategies $s$.
One desirable property of surrogate losses is \textit{Bayes-consistency}.
This property ensures that, in the asymptotic regime, minimising a surrogate loss $L_{surr}$ over the entire hypothesis class $\gQ$ (\textit{e.g.,} $f$ and $s$ in LtD) guarantees the minimisation of the original intractable loss function $L_{orig}$ over the same class \citep{DBLP:conf/alt/MaoM025}, i.e.:
\begin{equation}
    \begin{aligned}
    \lim_{n\to\infty}\mcE_{L_{surr}}(q_n)-\mcE^*_{L_{surr}}=0 \implies
\lim_{n\to\infty}\mcE_{L_{orig}}(q_n)-\mcE^*_{L_{orig}} =0
\label{eqn:bayes-consistency}
    \end{aligned}
\end{equation}
where $q_n$ is a sequence of hypotheses, each learned from $n$ samples.

In general, the property in Eqn.~\ref{eqn:bayes-consistency} is defined over the family of all measurable functions, which is impractical in real-world settings where learning is restricted to a limited hypothesis class.
To address this, a key property for surrogate losses is the \textit{realisable $\gQ$-consistency}.
Formally, a surrogate loss is \textit{relizable $\gQ$-consistent} if, whenever there exists a hypothesis $q^* \in \gQ$ such that $\mcE_{L_{orig}}(q^*) = 0$, we have that if $q' = \argmin_{q \in \gQ} \mcE_{L_{surr}}(q)$, then $\mcE_{L_{surr}}(q') = 0$.
In layman's terms, it means that if the “perfect” hypothesis (\textit{e.g.,} $(f,s)^*$ in LtD) exists in the given class, then minimising the surrogate loss will asymptotically recover it \citep{DBLP:conf/alt/MaoM025}.
An example of such a realizably consistent surrogate loss for LtD, proposed by \citet{DBLP:conf/nips/MadrasPZ18}, replaces the discontinuous deferral loss with a combination of cross-entropy terms:
\begin{align}
    (1-s(\mbx)) \cdot \ell_{CE}(f(\mbx),y) + s(\mbx) \cdot \ell_{CE}(f',y)
\end{align}
where $\ell_{CE}$ denotes the standard cross-entropy loss. 

\paragraph{Further related work.} LtD is an instance of hybrid decision making where the automated predictor learns when to defer predictions to a human expert. 
Due to the difficulties in directly optimizing Eqn.~(\ref{eq:LDEF}), consistent surrogate losses have been proposed to learn both the selection function and the predictor \cite{DBLP:conf/icml/MozannarS20,DBLP:conf/nips/OkatiDG21,DBLP:conf/icml/CharusaieMSS22,DBLP:conf/icml/VermaN22,DBLP:conf/aistats/MozannarLWSDS23,DBLP:conf/nips/CaoM0W023,DBLP:conf/aistats/LiuCZF024,DBLP:conf/icml/WeiC024}.
Recent works extend the LtD problem to account for multiple human experts~\citep{DBLP:conf/aistats/VermaBN23,DBLP:conf/nips/MaoMM023}; cases where the ML model is already given and not jointly trained~\citep{DBLP:conf/nips/MaoMM023}; and how to assess the causal effect of introducing LtD methods~\citep{Palomba25}.

Our work is closely related to both \citet{DBLP:conf/ijcai/WilderHK20} and \citet{charusaie2024defer}. Similarly, they consider the possibility of incorporating human expert feedback on the final task as additional input.
However, differently from us, they $(i)$ restrict the human feedback to the expert's prediction and $(ii)$ do not explicitly consider budget-aware querying of the expert feedback, which is a key aspect in real-life applications. 
Further, they lack any formal analysis of their learning approaches, while we provide new theoretical results for the optimal selection strategy (Theorem~\ref{thm:optimal_defer}) and realizable-consistency (Theorem~\ref{thm:real_consistency}).

\paragraph{Other human-AI frameworks.} Several works have focused on improving human-AI collaboration.
For instance, recent papers focus on how conformal prediction~\citep{DBLP:journals/jmlr/ShaferV08} can improve the joint decision-making of human-AI teams~\citep{DBLP:conf/icml/StraitouriR24,DBLP:conf/nips/ToniOTSR24}.
\citet{DBLP:conf/nips/AlurLLRSS23} develop a statistical framework to understand whether human experts can add value for a given prediction task, providing a data-driven procedure to test this hypothesis.
In \citet{DBLP:conf/nips/AlurRS24}, the authors identify if ML models are limited and measure if human expert feedback can improve them based on observed complementarities (ex-post).
We differ from their work as our focus is to learn models that know when to ask for expert feedback, while \citet{DBLP:conf/nips/AlurRS24} show how to test ex-post if expert feedback can improve predictions, without any learning involved. 

\begin{figure*}[t]
    \centering
    \includegraphics[width=\linewidth]{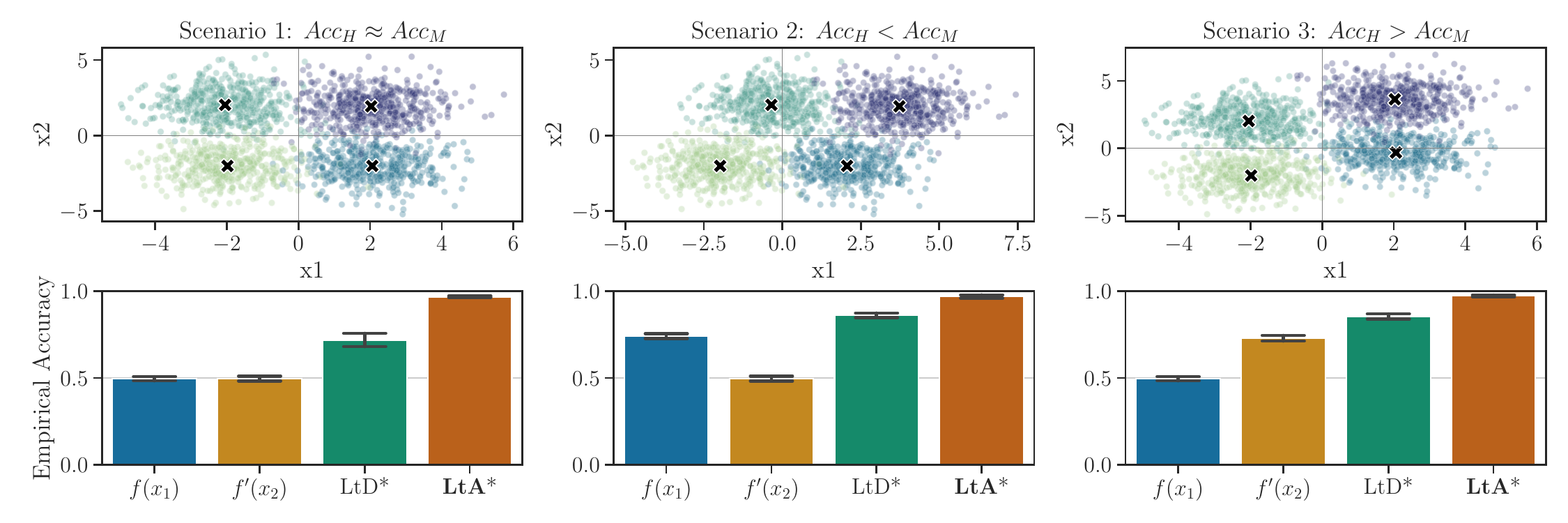}
    \caption{
    \textbf{Learning to Defer (LtD) can be suboptimal.}
    (Top) Synthetic classification task with two binary features $x_1,x_2 \in \{0,1\}$ and four classes $Y \in \{0,1,2,3\}$.
    We vary the \emph{informativeness} of each feature, \textit{i.e.,} how discriminative $x_1$ and $x_2$ are for predicting the true label, creating three scenarios, varying the expert ($Acc_H$) and machine ($Acc_M$) empirical accuracies on the test set. 
    (Bottom) For each scenario, neither the machine predictor $f(x_1)$ (\raisebox{0.5ex}{\fcolorbox[HTML]{FFFFFF}{0173B2}{\rule{0pt}{1pt}\rule{5pt}{0pt}}}) nor the human expert predictor $f'(x_2)$ (\raisebox{0.5ex}{\fcolorbox[HTML]{FFFFFF}{DE8F05}{\rule{0pt}{1pt}\rule{5pt}{0pt}}}) can perfectly recover $Y$ alone given their single feature.
    The oracle deferral strategy LtD* (\raisebox{0.5ex}{\fcolorbox[HTML]{FFFFFF}{029E73}{\rule{0pt}{1pt}\rule{5pt}{0pt}}}), which always chooses the correct prediction when possible, remains limited by this partial information.
    Only a strategy that integrates both signals, \textbf{LtA*} (\raisebox{0.5ex}{\fcolorbox[HTML]{FFFFFF}{D55E00}{\rule{0pt}{1pt}\rule{5pt}{0pt}}}), can achieve perfect classification.
    For each scenario, we report the standard deviation over five runs. 
    }
    \label{fig:simple-example-LtA}
\end{figure*} 

\section{On the Suboptimality of Learning to Defer}
\label{sec:suboptimal}

In the previous section, we have shown how LtD admits practical and consistent surrogate loss functions to minimise the objective in Eqn.~\ref{eq:LDEF} with finite sample guarantees.
Nevertheless, there exist data distributions where the optimal deferring policy under Learning to Defer (LtD) can be suboptimal.
Consider a stylized classification task with a discrete random variable $\mbX=(x_1,x_2)$, where $x_1,x_2  \in \{0,1\}$, and a label variable $Y \in \{0,1,2,3\}$. 
Each label $y \in Y$ corresponds deterministically to a unique combination of $x_1$ and $x_2$. 
Let $x_1$ represent a feature used by an automated decision support system (\textit{e.g.,} a complex function of vital signs), and $x_2$ a costly feature accessible only by processing additional information from a human expert (\textit{e.g.,} oral medical history).
Denote the machine’s prediction by $f(x_1)$, based solely on $x_1$, and the expert’s prediction by $f'(x_2)$ based solely on $x_2$.
Both agents are assumed to act as Bayes optimal predictors.
If the ML model can only observe $x_1$, we have $P\left(f(x_1) \mid x_1 = 0\right) = \nicefrac{1}{2}$ and $P\left(f(x_1) \mid x_1 = 1\right) = \nicefrac{1}{2}$ for $f(x_1) \in \{0,2\}$ and $f(x_1) \in \{1,3\}$, respectively.
Conversely, $P\left(f'(x_2) \mid x_2 = 0\right) = \nicefrac{1}{2}$ and $P\left(f'(x_2) \mid x_2 = 1\right) = \nicefrac{1}{2}$ for $f'(x_2) \in \{0,1\}$ and $f'(x_2) \in \{2,3\}$, respectively.
Given an instance $\mbx$, the expected accuracy  of an LtD system with strategy $s(x_1)$ is a convex combination of the expert's and the machine's accuracy:
\begin{equation}
    \begin{aligned}
        \mathbb{E}_{y}\left[s(x_1)\indicator{f(x_1)=y} + (1-s(x_1))\indicator{f'(x_2)=y}\right]
    \end{aligned}   
\end{equation}

In this setting, no selector $s(\cdot)$ can achieve an expected accuracy greater than $\nicefrac{1}{2}$.
The LtD system fails to exploit the complementary information available to the expert and the ML model,  \textit{i.e.,} $x_2$ in our example.
Now consider an alternative model that merges expert input into the classifier, \textit{e.g.,} the expert provides the collected feature $x_2$ to a unified model.
Let $g(x_1,x_2)$ denote the prediction of this combined system.
With access to both $x_1$ and $x_2$, the model can now perfectly predict the true label $Y$, achieving optimal accuracy.
Then, for any $s(x_1)>0$, the enriched system using $g(x_1, x_2)$ achieves an accuracy strictly greater than \nicefrac{1}{2}.
Figure~\ref{fig:simple-example-LtA} shows empirically the suboptimality of the oracle LtD, in three different scenarios, based on the base accuracy of the predictor and the human expert, given the features' informativeness.
See Appendix \ref{app:example_derivation} for the full derivation and a description of the setup.  

Nevertheless, obtaining online expert feedback, in the form of a prediction or additional features for each example, might be costly, and not always possible\footnote{
For example, consider a decision support system for the Emergency Department (ED) triage. 
ED physicians face frequent interruptions and a high workload \citep{chisholm2001work}.
%
%
Thus, the automated system must carefully optimise when to request additional expert feedback.
}.
Thus, in the next section, we focus on characterising the optimal selection strategy under \textit{budget}, balancing the need to elicit additional feedback from a human expert with maximising the system's accuracy. 

\section{Learning to Ask}
\label{sec:LtAformal}
We aim to extend Eqn.~(\ref{eq:LDEF}) by explicitly modelling $f'$ as a different function $g:\mcX\times\mcH \to \mcY$, where $\mcH\subseteq\mathbb{R}^d$ is the space of expert feedback, with $d>0$.
Intuitively, one can think of $\mcH$ as some other costly features to get beyond $\mcX$, including expert predictions for the task at hand (as in LtD), expert uncertainty predictions (\textit{e.g.,} how certain is the doctor about its diagnosis), extra features (\textit{e.g.,} some extra tests run by a physician), or some unstructured annotation (\textit{e.g.,} the whole medical report). 
We call $g$ the \textit{enriched predictor}, while we refer to $f$ as the \textit{standard predictor}. Notably, LtD is a specific setting of our approach,  \textit{i.e.,} whenever $\mcH=\mcY$ and $g(\mbx, h)=h$. 
Let us define the $L_{0-1}^{ask}$ loss as:
\begin{equation}
   \LZeroOnearg = \indicator{f(\mbx)\neq y}\indicator{s(\mbx)=0} + \indicator{g(x,h)\neq y}\indicator{s(\mbx)=1}
    \label{eq:LtASkOrigLoss}
\end{equation}
where $s(\mbx):\mcX\to \{0,1\}$ is a selection function, which determines which model will provide the final prediction,  \textit{i.e.,} when $s(\mbx)=0$ the standard model predicts, otherwise the enriched model predicts.
Since this loss is also intractable and hard to directly optimise, we consider a surrogate loss $L^{ask}$ defined as follows:
\begin{equation}
 \LAskarg = \ell^f\left(f(\mbx), y\right)(1-s(\mbx))+\ell^g\left(g(\mbx,h), y\right)s(\mbx), 
 \label{eq:LtA}
\end{equation}
where $\ell^f$ is a surrogate loss function for the standard predictor, and $\ell^g$ is a (possibly distinct) surrogate loss function for model $g$. Given a  distribution $P$ over $\mcX\times\mcY\times\mcH$, the goal of LtA is therefore the following:
\begin{equation}
    \min_{f,g,s\in \mcF\times\mcG\times\mcS} \mathbb{E}_{(\mbx, y, h)}\left[\LAskarg\right]
    \label{eq:LtA_min_problem}
\end{equation} 
To avoid over-reliance on the expert, we include a \textit{budget constraint} $\beta$ to Eqn.~\ref{eq:LtA_min_problem}, \textit{i.e.,} we require the standard model $f$ to cover $(1-\beta)\%$ of the instances, while querying the expert at most $\beta\%$ of times\footnote{Modeling the cost of expert with a budget constraint $\beta$ or with the parameters $\alpha$ and $\delta$ has a long story in the abstention literature. See, \textit{e.g.,}~\citet{DBLP:journals/jmlr/FrancPV23,DBLP:conf/aaai/RuggieriP25} for a more detailed discussion.}.
Thus, the minimization problem becomes:
\begin{equation}
    \min_{f,g,s\in \mcF\times\mcG\times\mcS} \mathbb{E}_{(\mbx, h, y)
    }\big[\LAskarg \big] \qquad
    \text{s.t.} \,\, \mathbb{E}_{\mbx
    }\left[s(\mbx)\right]\leq \beta\quad \quad\quad
    \label{eq:LtA_min_problem_coverage}
\end{equation}
Similarly to \citet{DBLP:conf/nips/OkatiDG21}, given any predictor $f\in\mcF$ and enriched predictor $g\in\mcG$, we can characterize the optimal selection function $s^*\in\mcS$ as follows:
\begin{theorem}
    Let $f\in\mcF$ be a standard predictor, and let $g\in\mcG$ be a fixed enriched predictor. Define $\Delta\mathbb{E}(\ell^f, \ell^g) = \mathbb{E}_{y\mid \mbx}[\ell^f\left(f(\mbx),y\right)]-\mathbb{E}_{y,h\mid\mbx}[\ell^g\left(g(\mbx,h), y\right)]$ as the difference between the expected conditional risk of $f$ and $g$ and assume that it is continuous.
    Then the optimal selection function $s^*$ for the minimization problem in Eqn.~(\ref{eq:LtA_min_problem_coverage}) can be written as:
    \begin{equation}
        s^*(\mbx) = \begin{cases}
            1 \quad \text{if} \quad \Delta\mathbb{E}(\ell^f, \ell^g) > \tau_\beta^*\\
            0 \quad \text{otherwise},
        \end{cases}
    \end{equation}
    where  $\tau_\beta^* = 
\inf_{\lambda}\{\lambda\geq0:  P(\mathbb{E}_{y\mid \mbx}
    \left[
        \ell^f\left( f(\mbx), y \right)
    \right] - \mathbb{E}_{y,h\mid \mbx}\left[
    \ell^g\left(
    g(\mbx, h), y
    \right)
    \right] >\lambda) \leq \beta \}$.
\label{thm:optimal_defer}
\end{theorem}

Theorem \ref{thm:optimal_defer} describes the optimal selection function under budget $\beta$: one queries the enriched predictor only if the difference between the standard predictor's expected loss and the expected loss of the enriched model is greater than the $(1-\beta)$-th quantile of such a difference.
See Appendix~\ref{app:proof-theorem-1} for the full proof.
In the next section, we will show how we can practically learn LtA methods via tailored surrogate functions, by also showing they are realizable and consistent.

\subsection{How to Practically Learn LtA Methods}
\label{sec:training-methods-lta}
Deploying an LtA method requires learning three models: the predictor ($f$), the enriched predictor ($g$), and the selection strategy ($s$).
We propose two different approaches: \textsc{LtA-Seq}, where the models are trained sequentially, and \textsc{LtA-Joint}, where all the models are trained jointly.
For the former, we exploit existing theoretical guarantees from the LtD literature \citep{DBLP:conf/icml/MozannarS20,DBLP:conf/alt/MaoM025}, while for the latter, we propose a novel realizable consistent surrogate loss (\textit{cf.} Theorem~\ref{thm:real_consistency}).

\paragraph{Sequential Training} (\Seq). We first propose to learn the $f$, $g$ and $s$ by dividing the learning into two steps: $(i)$ learning the model $g$ that requires additional human expert feedback, and then $(ii)$ learning $f$ and $s$ afterwards. 
This strategy is appealing because it leverages existing LtD theoretical guarantees to learn both the predictor $f$ and the selector $s$ as a single, unique model, as is commonly done in score-based LtD approaches~\citep{DBLP:conf/nips/MaoMM023}.
Indeed, once the model $g$ is fixed, we can use it to provide the expert predictions required in the LtD paradigm to train the defer model.
In this way, one can benefit from existing theoretical guarantees from the LtD literature.

\paragraph{Joint Training} (\Joint). Another option is to learn all three models jointly. However, existing LtD methods do not consider the case where the ``expert'' is jointly trained with the deferral model.
As done under the selector-predictor framework~\citep{DBLP:conf/nips/MaoMM023}, we exploit a surrogate loss $\ell^s$ to learn the selector in Equation \ref{eq:LtA}. Hence, we obtain the following loss:
\begin{equation}
     \LJAskarg = \ell^f\left(\tf_y(\mbx), y\right)\ell^s(-\ts(\mbx))+\ell^g\left(\tg_y(\mbx,h), y\right)\ell^s(\ts(\mbx)),  
     \label{eq:tildeLtA}
\end{equation}
where $\tf_y(\mbx):\mcX\to\mathbb{R}$ refers to the logit of the standard model $f$ associated to class $y$, $\tg_y(\mbx,h):\mcX\times\mcH\to\mathbb{R}$ is the logit of the enriched model $g$ associated to class $y$ and $\ts(\mbx):\mcX\to\mathbb{R}^+$ is a logit associated to a selection function, such that $s(\mbx)=\indicator{\ts(\mbx)>0}$.
Under some assumptions on the surrogate losses $\ell^f$,  $\ell^g$ and $\ell^s$, we show that the $\tilde{L}^{ask}$ surrogate loss (Equation \ref{eq:tildeLtA}) is $\mcF\times\mcG\times\mcS$-realizable consistent, as formalized in the next theorem:

\begin{theorem}
Assume that $\mcF$,$\mcG$ and $\mcS$ are closed under scaling, and consider the following surrogate losses such that
    \begin{itemize}
        \item[$(a)$] $\ell^f=\ell^g = \ell^{surr}$ is a comp-sum loss \citep{DBLP:conf/icml/MaoM023a} $\ell^{surr}:[0,1]\times\mcY \to [0,1]$ such that for a given $y$ is non-increasing in the first argument, $\ell^{surr}(\nicefrac{2}{3},y)>0$ and $\lim_{v\to 1}\ell^{surr}(v,y)=0$;
        \item[$(b)$] $\ell^s$ is a margin-based monotone loss such that $\lim_{v\to\infty}\ell^s(v) = 1$, $\lim_{v\to-\infty}\ell^s(v) = 0$; $\ell^s(-\tilde{s}(\mbx)) + \ell^s(\tilde{s}(\mbx))=1$;
    \end{itemize}
    then, the surrogate loss $\tilde{L}^{ask}$ (Eq. \ref{eq:tildeLtA}) is realizable $\mcF\times\mcG\times\mcS$-consistent with respect to $L^{ask}_{0-1}$.
    \label{thm:real_consistency}
\end{theorem}

Theorem \ref{thm:real_consistency} shows that considering surrogate losses that satisfy conditions $(a)$ and $(b)$ makes $\tilde{L}^{ask}$ a realizable  $\mcF\times\mcG\times\mcS$-consistent surrogate loss. Notice that assumption $(a)$ requires the loss $\ell^{surr}$ to take inputs in $[0,1]\times\mcY$. For a fixed value $y$, this can be easily done in practice by using the softmax of the logits, which we denote as $\smfy(\mbx)= \nicefrac{\exp{\tilde{f}_y(\mbx)}}{\sum_{y'\in\mcY}\exp \tilde{f}_{y'}(\mbx)}$ and $\smgy(\mbx,h)= \nicefrac{\exp{\tilde{g}_y(\mbx,h)}}{\sum_{y'\in\mcY}\exp \tilde{g}_{y'}(\mbx,h)}$.
A practical approach to define surrogate losses satisfying Theorem \ref{thm:real_consistency} is resorting to ${MAE}$ loss for both $\ell^f$ and $\ell^g$ and ${SIG}$ for $\ell^s$:
\begin{equation}
\ell^f_{MAE}(\smfy,y) = 1-\frac{\exp{\tilde{f}_y(\mbx)}}{\sum_{y'\in\mcY}\exp \tilde{f}_{y'}(\mbx)}
\label{eq:fmae}
\end{equation}
\begin{equation}
\ell^g_{MAE}(\smgy,y) = 1-\frac{\exp{\tilde{g}_y(\mbx,h)}}{\sum_{y'\in\mcY}\exp \tilde{g}_{y'}(\mbx,h)}
\label{eq:gmae}
\end{equation}
\begin{equation}
L_{SIG}^s(\tilde{s}(\mbx)) = \frac{1}{2}(1-\tanh{\tilde{s}(\mbx))} 
\label{eq:ssig}
\end{equation}
Indeed, it is immediate that these losses satisfy the requirements of Theorem 2.
%

\section{Experimental Evaluation}
\label{sec:experiments}

In this section, we evaluate empirically the effectiveness of LtA on simulated decision-making tasks with synthetic and real expert data, under different feedback strategies.
We aim to address the following research questions:
\begin{itemize}
    \item [\textbf{Q1:}] How do LtA strategies compare to LtD in terms of \textit{accuracy} and \textit{human-AI complementarity}?
    \item [\textbf{Q2:}] To what extent does the type of expert feedback influence performance and complementarity?
    \item [\textbf{Q3:}] What is the effect of the defer cost $\delta$ on mitigating underfitting and balancing performance?

\end{itemize}

\paragraph{Datasets and models.}
We first construct a synthetic multiclass classification task (\synth) with four features and four labels, where all features are informative.
To simulate a human expert, we train a decision tree $f'(\mbx)$ using only two features, while the machine model is a two-layer perceptron trained on the remaining two.
This setup reflects an expert having access to complementary information unavailable to the predictor (similar to the example in Section~\ref{sec:suboptimal}).
Second, we consider the NIH Google Chest X-ray dataset~\citep{chestxray} (\chx), which provides $\approx4.4k$ anonymized chest X-ray images with expert annotations.
Each image is evaluated by at least three physicians for four conditions ($c_i$): nodule/mass, airspace opacity, pneumothorax, and fracture.
For each condition $i \in [4]$, images are assigned a ground-truth binary label ($c_i=1$ if present, $c_i=0$ otherwise).
We define an overall binary label $Y \in \{0,1\}$, where $Y=1$ denotes a healthy case ($c_i=0$ for all $i$) and $Y=0$ otherwise.
For each condition $c_i$, we estimate the \textit{expert consensus} $p_i$ as the proportion of physicians reporting the condition $p_i = \frac{1}{N} \sum_{j \in N}{c_{ij}}$, where $c_{ij}$ is expert $j$’s binary annotation and $N$ is the number of annotators.
To simulate the expert prediction $Y_H$, we sample condition labels as $c_i \sim \text{Bernoulli}(p_i)$ and assign $Y_H=1$ if no condition is detected ($c_i=0$ for all $i$). This procedure reflects the variability in expert agreement when determining patient health status.
The machine is a pre-trained deep neural network DenseNet as provided by \citep{Cohen2022xrv}\footnote{
In practice, for both LtA and LtD, we fine-tune the DenseNet model on the \chx~dataset during training, following standard practice in the literature.
}.
Lastly, we consider a deep neural network as an enriched model $g$, where we inject the human expert feedback using the FiLM architecture~\citep{DBLP:conf/aaai/PerezSVDC18}, a well-established approach for visual reasoning.

\paragraph{Expert feedback.} 
We define two alternative feedback strategies: (i) \texttt{LtD-Feedback}, where $h=f’(\mbx)$ and $h=Y_H$ (for \synth{} and \chx, respectively) simulating the expert providing only the predicted health label to the enriched model; and (ii) $\texttt{Unc-Feedback}$, where $h=(p_1, p_2, p_3, p_4)$ (only for \chx) simulating the expert providing consensus probabilities for each condition, reflecting the expert's uncertainty about their presence in the image.
In LtD, we consider only the \texttt{LtD-Feedback} since the expert is required to provide their prediction directly.

\paragraph{Experimental setup.} For both datasets, we split the data into training, calibration, and test sets in a 70:10:20 ratio.
The training set is used to train both standard and enriched models under the LtD, \Seq and \Joint strategies.
For the standard LtD and \Seq settings, we adopt the realizable consistent loss of \citet{DBLP:conf/aistats/MozannarLWSDS23} and \citet{DBLP:conf/nips/MaoM024b}, which achieves state-of-the-art performance in LtD.
For \Joint, we train the machine model $f$, the enriched model $g$, and the selector using the losses: Eq.~\ref{eq:fmae}, Eq.~\ref{eq:gmae}, and Eq.~\ref{eq:ssig}, respectively.
At test time, following standard practice~\citep{DBLP:conf/aistats/MozannarLWSDS23,Palomba25}, we use the calibration set to estimate the deferral threshold $\tau_\beta$ for each coverage level $\beta \in \{0.0, \dots, 1.0\}$.
We then compute the empirical accuracy of each method on the test set across thresholds.
For comparison, we also measure the accuracy of the (simulated) expert and the model when acting alone, to assess \textit{complementarity}, that is, whether human-machine teams outperform both individual components \citep{steyvers2024three}.
Lastly, we repeat the experiments five times, by sampling each time a different training, calibration and test set.
Additional details are available in Appendix~\ref{app:further-experimental-details}.

\begin{figure*}[t]
    \centering
    \begin{subfigure}{.32\linewidth}        
    \includegraphics[scale=.37]{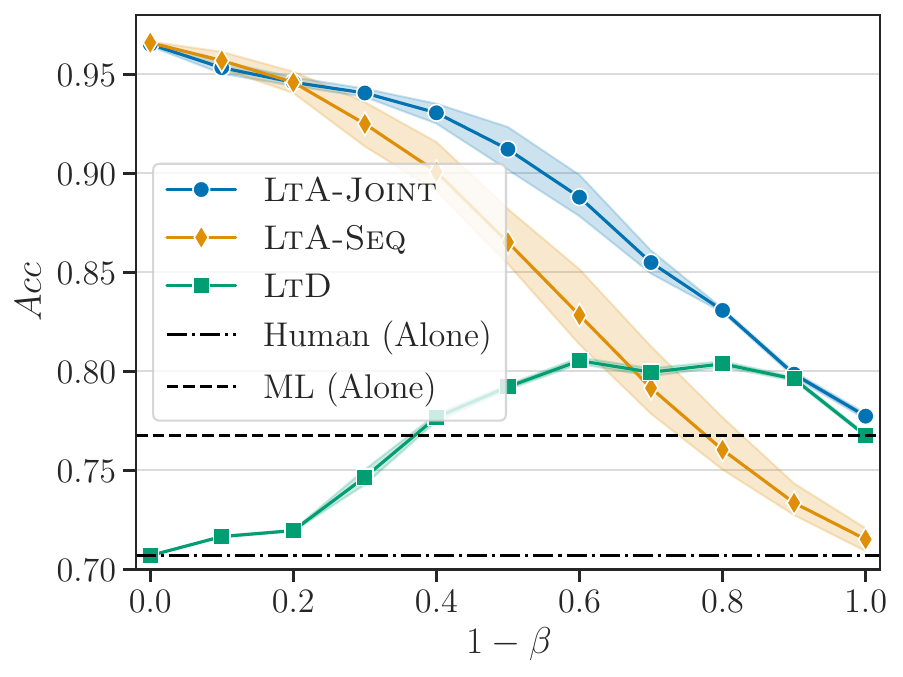}
    \caption{\synth{}}
    \label{fig:acc_SynthZeroCost}
    \end{subfigure}
    \begin{subfigure}{.33\linewidth}  
    \centering
    \includegraphics[scale=.37]{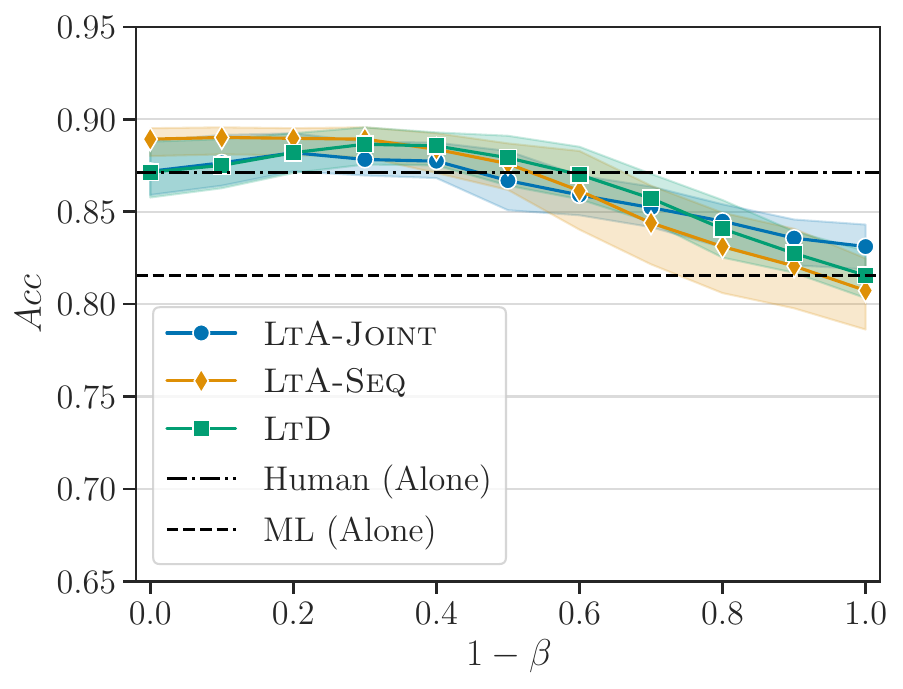}
    \caption{\chx{} (\texttt{LtD-Feedback})}
    \label{fig:acc_CHXLtDZeroCost}
    \end{subfigure}
    \begin{subfigure}{.33\linewidth}        
    \includegraphics[scale=.37]{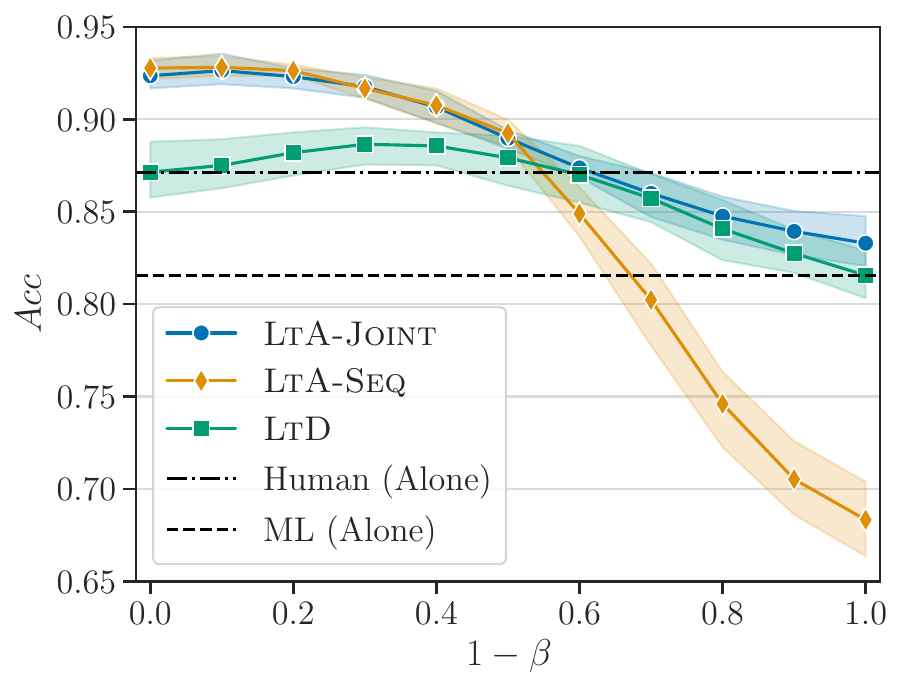}
    \caption{\chx{} (\texttt{Unc-Feedback})}
    \label{fig:acc_CHXUncZeroCost}
    \end{subfigure}
    
    \caption{
    \textbf{Empirical accuracy ($Acc$) at various coverage levels ($1-\beta$), on both synthetic and real classification tasks, and different expert feedback.}
    (\ref{fig:acc_SynthZeroCost}) Results for \synth{}; (\ref{fig:acc_CHXLtDZeroCost}) Results for \chx{} when using standard LtD feedback; (\ref{fig:acc_CHXUncZeroCost}) Results for \chx{} when using uncertainty feedback to train LtA methods. In Fig.~\ref{fig:acc_CHXUncZeroCost}, we also include the \textsc{LtD} baseline.
    We also report the empirical accuracy of the expert and machine predictions alone as dotted lines. 
    Lastly, we report the standard deviation over five runs as shaded areas.
    }
    \label{fig:Q1ANDQ2}
\end{figure*}

\subsection{Experimental results}

\paragraph{Q1: \Joint and \Seq outperforms traditional LtD, with \Seq sensitive to underfitting.}
We first compare our approaches, \Seq and \Joint, against standard LtD methods on the synthetic classification task (\synth).
Allowing for more human feedback yields a clear advantage for LtA: both \Seq and \Joint achieve substantial gains over LtD for $1-\beta < .50$.
In particular, \Joint consistently outperforms \textsc{LtD} across all coverage levels, while \Seq attains lower accuracy than \textsc{LtD} for $1-\beta \geq .70$.
Further, by exploiting feedback more effectively, \textit{LtA achieves stronger complementarity than LtD}, surpassing the empirical accuracy of both the simulated expert and machine predicting alone.

\begin{figure*}[t]
    \centering
    \begin{subfigure}{.32\linewidth}        
    \includegraphics[scale=.37]{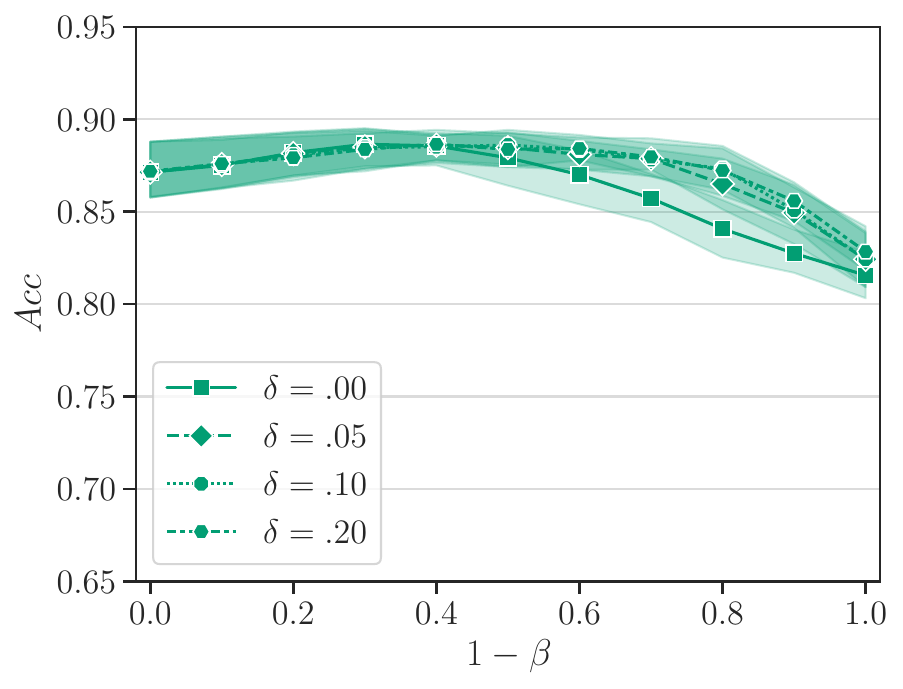}
    \caption{\textsc{LtD} varying $\delta$}
    \label{fig:deltaLtD}
    \end{subfigure}
    \begin{subfigure}{.32\linewidth}        
    \includegraphics[scale=.37]{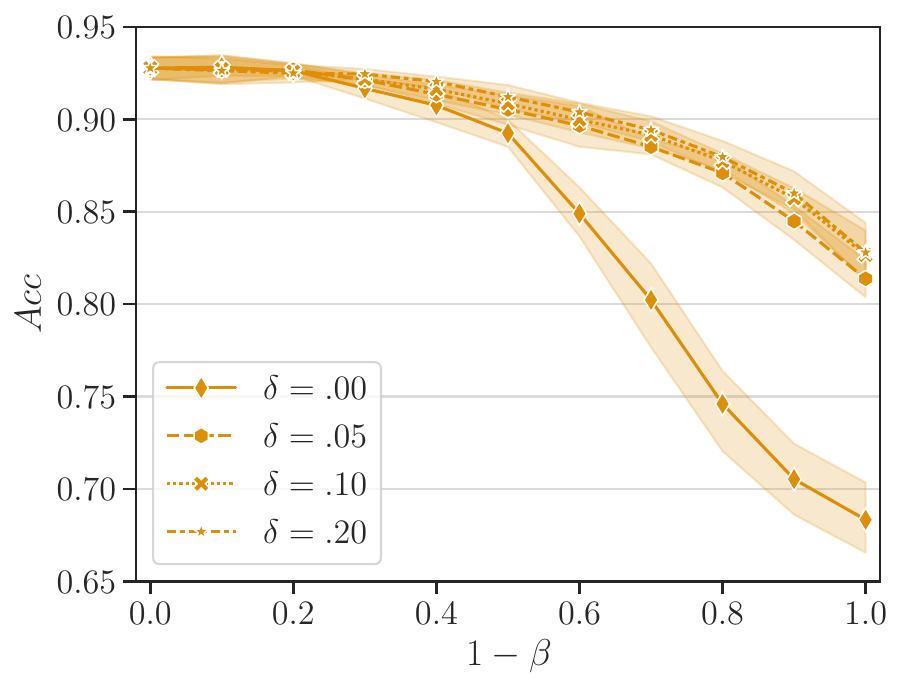}
    \caption{\Seq varying $\delta$}
    \label{fig:deltaSeq}
    \end{subfigure}
    \begin{subfigure}{.32\linewidth}        
    \includegraphics[scale=.37]{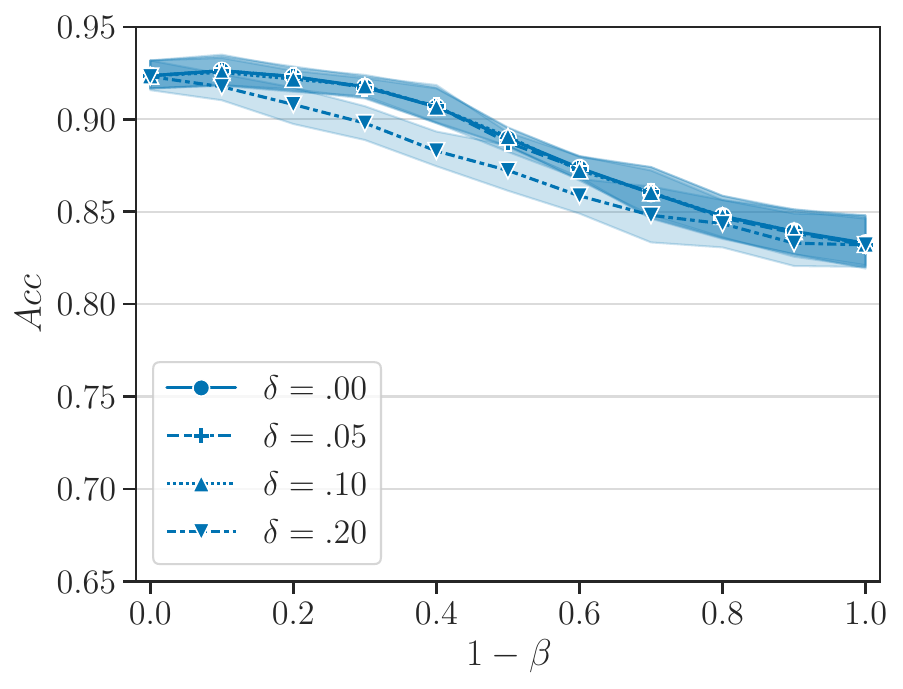}
    \caption{\Joint varying $\delta$}
    \label{fig:deltaJoint}
    \end{subfigure}
    
    \caption{\textbf{Empirical accuracy ($Acc$) on \chx{} at various coverage levels ($1-\beta$), using different defer costs $\delta \in \{0.0, \ldots, 1.0\}$.} (\ref{fig:deltaLtD}) Results for \textsc{LtD} (\ref{fig:deltaSeq}, \ref{fig:deltaJoint}) Results for \Seq and \Joint, respectively, with uncertainty feedback.
    We report the standard deviation over five runs as shaded areas.
    }
    \label{fig:Q3}
\end{figure*}

The performance gap between \Seq and \Joint at high coverage can be traced to their training procedures. As described in Sec.~\ref{sec:training-methods-lta}, \Seq first trains the enriched model $g$ without penalizing expert queries ($\delta=0$).
Since $g$ outperforms the simulated expert by roughly $20\% $ at $1-\beta=0$, \Seq tends to over-rely on $g$, which in turn leads to underfitting of the other model heads, a known issue in LtD~\citep{DBLP:conf/aistats/LiuCZF024}.
Nevertheless, we show in Fig.~\ref{fig:Q3} that simply increasing the defer cost mitigates this effect and enables \Seq to approach the performance of \Joint on real expert data.
In contrast, \Joint naturally avoids this over-reliance: its joint training procedure acts as a regularizer, limiting expert queries and leveraging feedback only when beneficial.

\paragraph{Q2: On real data, the performance of LtA is equivalent to LtD with hard predictions, but increases significantly with richer forms of feedback.}
We next evaluate alternative feedback strategies (\texttt{LtD-Feedback} and \texttt{Unc-Feedback}) using real-world expert annotations from \chx{}.
When relying on traditional LtD feedback ( \textit{i.e.,} hard predictions), empirical accuracy remains largely unchanged: \Joint and \Seq perform on par with \textsc{LtD}, and overall complementarity is limited, with all methods barely exceeding the accuracy of the expert alone (Fig.~\ref{fig:acc_CHXLtDZeroCost}).
In contrast, providing more informative feedback through consensus probabilities (\texttt{Unc-Feedback}) yields clear benefits for LtA.
Both \Seq and \Joint outperform \textsc{LtD}, achieving up to a $\sim 5\%$ improvement at $1-\beta=.10$.

By leveraging richer feedback, these strategies achieve \textit{stronger complementarity}, surpassing the performance of both human experts and the machine alone for $1-\beta \leq .05$.
As observed previously, \Seq continues to exhibit underfitting for $1-\beta \geq .50$ when using \texttt{Unc-Feedback}, which is expected given the increased complexity of the feedback compared to hard predictions.

\paragraph{Q3: Increasing the defer cost reduces \Seq underfitting toward \Joint performance.}
Lastly, we investigate the use of the defer cost $\delta$ during training as a way to mitigate the underfitting observed in traditional LtD methods~\citep{DBLP:conf/aistats/MozannarLWSDS23,DBLP:conf/aistats/LiuCZF024}.
Figure~\ref{fig:Q3} reports results on the real-world \chx classification task for varying values of $\delta$ across all methods.
For \textsc{LtD}, introducing a defer cost yields moderate gains at higher coverage levels: for $1-\beta \in \{.80, .90\}$, setting $\delta=0.2$ improves accuracy by roughly $3\%$ (Fig.~\ref{fig:deltaLtD}).
More notably, \Seq shows \textit{substantial improvements} (Fig.~\ref{fig:deltaSeq}), with significant gains for $1-\beta > .50$ and a maximum increase of about $16\%$ at $1-\beta = .90$ when $\delta=0.2$.
These results indicate that incorporating a defer cost reduces overreliance on the expert component, effectively mitigating underfitting.

While the effect is modest for LtD, which underperforms compared to LtA with informative feedback, in the case of \Seq, a defer cost enables performance comparable to \Joint, further supporting the effectiveness of LtA.
Interestingly, for \Joint (Fig.~\ref{fig:deltaJoint}), increasing $\delta$ \textit{slightly degrades} performance.
We attribute this to the fact that penalizing expert feedback makes it more challenging for the joint strategy to optimize the enriched model effectively.

\section{Conclusions}

We introduced Learning to Ask (LtA), a novel framework for human–AI collaboration that learns when to query a human expert and integrates this feedback into prediction, rather than simply deferring as in Learning to Defer (LtD).
We first showed that there exist data distributions where LtD is suboptimal.
We then derived the optimal selection strategy, including the case of budgeted expert queries, and proposed two practical training approaches: a sequential regime and a joint regime with a novel realizable-consistent surrogate loss.
Lastly, we have empirically shown that LtA consistently matches or outperforms LtD, while offering greater flexibility in incorporating diverse forms of expert feedback.

\paragraph{Limitations and future work.}
We conclude by discussing some limitations that open up interesting avenues for future work.
Some theoretical aspects remain open, \textit{e.g.,} how to design surrogate losses that benefit from $\mcF\times\mcG\times\mcS$-consistency bounds.
In Theorem~\ref{thm:real_consistency}, we show how to characterize a $\mcF\times\mcG\times\mcS$-realizable consistent surrogate loss for LtA.
However, although we show that by minimizing the surrogate loss, we can retrieve the optimal $(f,g,s)^* \in \mcF\times\mcG\times\mcS$, we still lack Bayes-consistency. This limitation highlights the need for more refined analyses, such as exploring $\gQ$-consistency bounds~\citep{DBLP:conf/alt/MaoM025}.
Another important open question is understanding which type of human feedback is most useful for a given instance. While our current formulation relies on a \textit{fixed query structure}, real-world applications may require adapting the form of queries, ranging from simple predictions to richer annotations (\textit{e.g.,} uncertainty estimates, structured explanations, etc.). Developing mechanisms that not only decide \textit{when to defer} but also \textit{what to ask} represents a promising research direction, enabling more effective use of human expertise.

\section*{Acknowledgements}

This work was partially supported by the following projects: Horizon Europe Programme, grants \#10112\-0237-ELIAS and \#101120763-TANGO.
Funded by the European Union. Views and opinions expressed are however those of the author(s) only and do not necessarily reflect those of the European Union or the European Health and Digital Executive Agency (HaDEA). Neither the European Union nor the granting authority can be held responsible for them.
This work was also supported by Ministero delle Imprese e del Made in Italy (IPCEI Cloud DM 27 giugno 2022 – IPCEI-CL-0000007), PNRR-M4C2-Investimento 1.3, Partenariato Esteso PE00000013-“FAIR-Future Artificial Intelligence Research”, funded by the European Commission under the NextGeneration EU programme, and SoBigData.it, receiving funding from European Union – NextGenerationEU – National Recovery and Resilience Plan (Piano Nazionale di Ripresa e Resilienza, PNRR) – Project: “SoBigData.it – Strengthening the Italian RI for Social Mining and Big Data Analytics” – Prot. IR0000013 – Avviso n. 3264 del 28/12/2021.

\bibliography{biblio}
\bibliographystyle{plainnat}

\clearpage
\appendix
\thispagestyle{empty}
\onecolumn

\section{Proofs}

\subsection{Full derivation for the example}
\label{app:example_derivation}

We provide here the full derivations for the example of the suboptimality of Learning to Defer (LtD).

First, let us start by considering the expected accuracy of an LtD system.
Given a fixed instance $\mbx = (x_1, x_2)$, let us notice that we can easily decouple the expectation into two terms corresponding to (i) the expected accuracy of the human expert and (ii) the expected accuracy of the machine. 
\begin{align*}
    Acc\left(f,f',s(x_1)\right) & =\mathbb{E}_{y}\left[s(x_1)\indicator{f(x_1)=y}
    + (1-s(x_1))\indicator{f'(x_2)=y}\right]\\
    &= \sum_{y \in Y} P(Y=y \mid x_1, x_2)\left[s(x_1)\indicator{f(x_1)=y}
    + (1-s(x_1))\indicator{f'(x_2)=y}\right] \\
    &= s(x_1)\sum_{y \in Y} P(Y=y \mid x_1, x_2)\indicator{f(x_1)=y} \\
    & \qquad\qquad+ (1-s(x_1))\sum_{y \in Y} P(Y=y \mid x_1, x_2)\indicator{f'(x_2)=y}\\
    &=s(x_1)\mathbb{E}_{y}\left[\indicator{f(x_1)=y}\right] + (1-s(x_1))\mathbb{E}_{y}\left[\indicator{f'(x_2)=y}\right]
\end{align*}
Then, we can examine the two expectations separately.
Recall that, in our example, once we observe $x_1$ and $x_2$ jointly, we can deterministically assign the correct label.
Thus, we have $P(Y=y \mid x_1, x_2) = 1$ for each specific combination of $(y,x_1,x_2)$, and zero otherwise.
See Table~\ref{app:table:syn_example} for the complete data distribution. 
\begin{table}[b]
\centering
\caption{Distributions of the labels' probability, for each combination of features $\mbx = (x_1, x_2)$, for the synthetic classification task, described in Section~\ref{sec:suboptimal}.}
\label{app:table:syn_example}
\begin{tabular}{cccccc}
\toprule
\multicolumn{2}{c}{\multirow{2}{*}{$x_1$ $x_2$}} & \multicolumn{4}{c}{$P(Y=y \mid x_1, x_2)$} \\ \cmidrule(l){3-6} 
\multicolumn{2}{c}{}                             & $y=0$     & $y=1$    & $y=2$    & $y=3$    \\ \midrule
0                       & 0                      & 1         & 0        & 0        & 0        \\
0                       & 1                      & 0         & 1        & 0        & 0        \\
1                       & 0                      & 0         & 0        & 1        & 0        \\
1                       & 1                      & 0         & 0        & 0        & 1        \\ \bottomrule
\end{tabular}
\end{table}
Therefore, given a fixed instance $\mbx$, the expected accuracy of the human expert or the machine is simply:
\begin{align*}
 \mathbb{E}_{y}\left[\indicator{f(x_1)=y}\right] &= \sum_{y \in Y} P(Y=y \mid x_1, x_2) P(f(x_1)=y \mid x_1) \\
 &= 0\cdot 0 + 0\cdot 0 + 0 \cdot \frac{1}{2} + 1 \cdot \frac{1}{2} =  \frac{1}{2}
\end{align*}
Since $s(x_1) \in (0,1)$, then the overall expected accuracy $Acc\left(f,f',s(x_1)\right)$ is always $\nicefrac{1}{2}$.
If instead of deferring to a human expert, we were to use a Bayes-optimal enriched predictor, combining the machine signal and the expert-provided feature $x_2$, then we would obtain a stricter, higher accuracy.
\begin{align*}
 \mathbb{E}_{y}\left[\indicator{g(x_1,x_2)=y}\right] &= \sum_{y \in Y} P(Y=y \mid x_1, x_2) P(g(x_1,x_2)=y \mid x_1,x_2) \\
 &= 0\cdot 0 + 0\cdot 0 + 0 \cdot 0 + 1 \cdot 1 = 1
\end{align*}
Indeed, for any $s(x_1) > 0$ we would have:
\[
s(x_1)\mathbb{E}_{y,f}\left[\indicator{f(x_1)=y}\right] + (1-s(x_1))\mathbb{E}_{y,f'}\left[\indicator{g(x_1, x_2)=y}\right] = s(x_1)\frac{1}{2} + (1-s(x_1))1 \geq \frac{1}{2}
\]

\subsection{Expertimental Setup for Figure~\ref{fig:simple-example-LtA}}

We construct a synthetic classification task with four classes $Y \in \{0,1,2,3\}$, each defined by a pair of binary latent features $(b_1,b_2) \in \{0,1\}^2$. Each class is mapped to two observed features $(x_1,x_2)$, with centroids placed at $(\pm \text{sep}, \pm \text{sep})$ and perturbed with Gaussian noise. To control feature informativeness, we optionally shift centroids along the $x_1$ (machine) or $x_2$ (expert) axis: increasing separation along $x_1$ makes the machine more accurate, and analogously for the expert with $x_2$.

We generate 4000 instances, splitting them into training and test sets (60:40). Logistic regression models are trained using only $x_1$ (machine predictor $f(x_1)$), only $x_2$ (expert predictor $f'(x_2)$), or both features (enriched predictor $g(x_1,x_2)$). We then evaluate two oracle strategies: Learning to Defer (LtD), which dynamically chooses between machine and expert, and Learning to Ask (LtA), which optimally decides when to query expert feedback.
In this setting, expert feedback corresponds directly to the expert feature $x_2$.
Final results are reported in Fig.~\ref{fig:simple-example-LtA}.

\subsection{Proof of Theorem 1}
\label{app:proof-theorem-1}
\begin{proof}
We start by checking the unconstrained case, i.e., our goal is to minimize the following:
\begin{equation}
    \min_{s\in\mcS} \mathbb{E}_{(\mbx, y,h)}\left[\left(1-s(\mbx)\right)\ell^f\left(f(\mbx),y\right)+s(\mbx)\ell^g\left(g(\mbx, h),y\right)\right]
\end{equation}.
The objective can be written as
\begin{multline*}
    \mathbb{E}_{(\mbx, y, h)}\left[
    \left(1-s(\mbx)\right)\ell^f\left(f(\mbx),y\right)
    +s(\mbx)\ell^g\left(g(\mbx, h),y\right)
    \right] =\\
    \mathbb{E}_{\mbx}\left[
    \left(
    1 - s(\mbx)
    \right)
    \mathbb{E}_{y\mid \mbx}
    \left[
        \ell^f\left( f(\mbx), y \right)
    \right]
    +
    s(\mbx)
    \mathbb{E}_{y,h\mid \mbx}\left[
    \ell^g\left(
    g(\mbx, h), y
    \right)
    \right]
    \right] =\\
    \mathbb{E}_{\mbx}\left[
    \mathbb{E}_{y\mid \mbx}
    \left[
        \ell^f\left( f(\mbx), y \right)
    \right]
    \right]+
\mathbb{E}_{\mbx}\left[
s(\mbx) 
\left(
\mathbb{E}_{y,h\mid \mbx}\left[
    \ell^g\left(
    g(\mbx, h), y
    \right)
    \right] -
    \mathbb{E}_{y\mid \mbx}
    \left[
        \ell^f\left( f(\mbx), y \right)
    \right]
\right)
\right]
\end{multline*}
Notice that since the first term of the problem does not depend on $s$, we are solving the following optimization problem:
\begin{equation}
    \min_{s\in[0,1] \forall \mbx \in \mcX} \mathbb{E}_{\mbx}\left[
s(\mbx) 
\left(
\mathbb{E}_{y,h\mid \mbx}\left[
    \ell^g\left(
    g(\mbx, h), y
    \right)
    \right] -
    \mathbb{E}_{y\mid \mbx}
    \left[
        \ell^f\left( f(\mbx), y \right)
    \right]
    \right)
    \right]
\end{equation}
This is a linear problem, and it decouples with respect to $\mbx$. Hence, for every $\mbx\in\mcX$, it is immediate to check that the optimal selector is:
\begin{equation}
    s^*(\mbx) = \begin{cases}
        1 \quad \text{if} \quad  
    \mathbb{E}_{y\mid \mbx}
    \left[
        \ell^f\left( f(\mbx), y \right)
    \right] - \mathbb{E}_{y,h\mid \mbx}\left[
    \ell^g\left(
    g(\mbx, h), y
    \right)
    \right] >0 \\
    0 \quad \text{otherwise}
    \end{cases}
\end{equation}

Now, let us consider the case where we have a budget constraint, i.e., the problem becomes:
\begin{equation}
    \min_{s\in[0,1] \forall \mbx \in \mcX} \mathbb{E}_{(\mbx, h, y)}\left[\left(1-s(\mbx)\right)\ell^f\left(f(\mbx),y\right)+s(\mbx)\ell^g\left(g(\mbx, h),y\right)\right] \quad \text{s.t.} \quad\mathbb{E}_{\mbX}\left[s(\mbx)\right]\leq \beta
\end{equation}.

We can now consider the dual formulation, i.e.,

\begin{equation}
        \max_{\lambda \geq 0}\min_{s\in[0,1] \forall \mbx \in \mcX} \mathbb{E}_{(\mbx, h, y)}\left[\left(1-s(\mbx)\right)\ell^f\left(f(\mbx),y\right)+s(\mbx)\ell^g\left(g(\mbx, h),y\right)\right] + \lambda\mathbb{E}_{\mbx}\left[s(\mbx)-\beta\right],
\end{equation}

where $\lambda$ is the Lagrangian multiplier for the budget constraint.
The inner argument can be solved similarly to the unconstrained case, leading us to the following optimal selector:

\begin{equation}
    s^*_\lambda(\mbx) = \begin{cases}
        1 \quad \text{if} \quad  
    \mathbb{E}_{y\mid \mbx}
    \left[
        \ell^f\left( f(\mbx), y \right)
    \right] - \mathbb{E}_{y,h\mid \mbx}\left[
    \ell^g\left(
    g(\mbx, h), y
    \right)
    \right] >\lambda \\
    0 \quad \text{otherwise}
    \end{cases}.
\end{equation}
Then, notice that since:
\begin{equation*}
    s^*_{\lambda}(\mbx) = \indicator{\mathbb{E}_{y\mid \mbx}
    \left[
        \ell^f\left( f(\mbx), y \right)
    \right] - \mathbb{E}_{y,h\mid \mbx}\left[
    \ell^g\left(
    g(\mbx, h), y
    \right)
    \right] >\lambda},
\end{equation*}
we can write that
\begin{multline*}
\mathbb{E}_\mbx[s^*_\lambda(\mbx)] = \mathbb{E}_\mbx[\indicator{\mathbb{E}_{y\mid \mbx}
    \left[
        \ell^f\left( f(\mbx), y \right)
    \right] - \mathbb{E}_{y,h\mid \mbx}\left[
    \ell^g\left(
    g(\mbx, h), y
    \right)
    \right] >\lambda}] = \\
    P_\mbx\left(\mathbb{E}_{y\mid \mbx}
    \left[
        \ell^f\left( f(\mbx), y \right)
    \right] - \mathbb{E}_{y,h\mid \mbx}\left[
    \ell^g\left(
    g(\mbx, h), y
    \right)
    \right] >\lambda\right)= \\1-P_\mbx(\mathbb{E}_{y\mid \mbx}
    \left[
        \ell^f\left( f(\mbx), y \right)
    \right] - \mathbb{E}_{y,h\mid \mbx}\left[
    \ell^g\left(
    g(\mbx, h), y
    \right)
    \right] \leq\lambda);
\end{multline*}

then at the optimum level $\lambda^*$ must satisfy the constraint $\mathbb{E}_\mbx[s(\mbx)]\leq \beta$ with equality, i.e., $\lambda^*$ satisfies
$P_\mbx(\mathbb{E}_{y\mid \mbx}
    \left[
        \ell^f\left( f(\mbx), y \right)
    \right] - \mathbb{E}_{y,h\mid \mbx}\left[
    \ell^g\left(
    g(\mbx, h), y
    \right)
    \right] \leq\lambda^*) = 1-\beta$.

Due to the continuity assumption,  $\lambda^*$ is the $(1-\beta)$-quantile for the expected difference $\mathbb{E}_{y\mid \mbx}
    \left[
        \ell^f\left( f(\mbx), y \right)
    \right] - \mathbb{E}_{y,h\mid \mbx}\left[
    \ell^g\left(
    g(\mbx, h), y
    \right)
    \right]$.
\end{proof}

\subsection{Proof of Theorem 2}
\label{app:proof-theorem-2}
To show Theorem 2 holds, we follow a similar approach to \citet{DBLP:conf/nips/MaoM024b}: first, we show that our surrogate loss upper bounds $L^{ask}_{0-1}$, then we show that if we assume realizability, our loss converges to zero.
Let us denote (with a slight abuse of notation) as $\ell^{surr}(v)$ the loss obtained when considering a logit whose softmax is $v$, i.e.,  $\ell^{surr}(v) =\ell^{surr}(v,y)$.
We first show that for every $\mbx, h, y$ and every $f,g,s \in \mcF\times\mcG\times\mcS$ it holds that:
\begin{equation}
    \ \LZeroOnearg \leq \frac{\LJAskarg}{\commonfrac},
\end{equation}
We consider six cases here:\\
\textbf{Case 1:} both predictors are wrong, i.e., 
$f(\mbx)\neq y$ and $g(\mbx,h)\neq y$. Then $L_{0-1}^{ask}(\cdot) = 1$.
Moreover, one can now notice that for $\tilde{L}^{ask}(\cdot)$ it holds that:
\begin{multline}
    \LJAskarg = \left(1-\ell^s\left(\tilde{s}(\mbx)\right)\right) \ell^f\left(\smfy,y\right)+\ell^s\left(\tilde{s}(\mbx)\right) \ell^g\left(\smgy, y\right) = \\
    \left(1-\ell^s\left(\tilde{s}(\mbx)\right)\right)\SL\left(\smfy,y\right)+\ell^s\left(\tilde{s}(\mbx)\right) \SL\left(\smgy, y\right) \geq \\
    \left(1-\ell^s\left(\tilde{s}(\mbx)\right)\right) \SL\left(\nicefrac{1}{2}, y\right)+\ell^s\left(\tilde{s}(\mbx)\right) \SL\left(\nicefrac{1}{2},y\right)  \geq \\
    \left(1-\ell^s\left(\tilde{s}(\mbx)\right)\right) \SL\left(\nicefrac{2}{3}, y\right)+\ell^s\left(\tilde{s}(\mbx)\right) \SL\left(\nicefrac{2}{3},y\right)
\end{multline}

where the first qeuality holds by assumption $(b)$, inequality holds because if $f(\mbx)\neq y$, then $\sigma(\tilde{f}_y)<\nicefrac{1}{2}$, as according to the Bayes Rule the prediction is such that $f(\mbx)\in\argmax_{y\in\mcY}\tilde{f}_y(\mbx)$ (similar reasoning holds for $g(\mbx, y)$). Hence, both softmax values for $f$ and $g$ are such that $\smfy \neq \max_{k\in\mid\mcY\mid}\sigma\left( \tf_k(\mbx)\right)$ and $\smgy(\mbx,h) \neq \max_{k\in\mid\mcY\mid}\sigma\left(\tg_k(\mbx,h)\right)$. This observation implies that $\smfy(\mbx) < \nicefrac{1}{2}$ and $\smgy(\mbx,h) < \nicefrac{1}{2}$. Then the second inequality holds since $\SL$ is non-increasing. Now let us divide both parts by the positive quantity $\commonfrac$:

\begin{multline}
   \frac{\LJAskarg}{\commonfrac} \geq
    \frac{\left(1-\ell^s\left(\tilde{s}(\mbx)\right)\right) \SL(\nicefrac{2}{3}, y)}{\commonfrac}+\frac{\ell^s\left(\tilde{s}(\mbx)\right) \SL(\nicefrac{2}{3},y)}{\commonfrac} \geq \\
    \frac{\left(1-\ell^s\left(\tilde{s}(\mbx)\right)\right) \SL(\nicefrac{2}{3}, y)}{\SL(\nicefrac{2}{3})}+\frac{\ell^s\left(\tilde{s}(\mbx)\right) \SL(\nicefrac{2}{3},y)}{\SL(\nicefrac{2}{3})} = 1= \LZeroOnearg
\end{multline}

where the second inequality holds as $\SL(\nicefrac{2}{3}) > \nicefrac{1}{2}\SL(\nicefrac{2}{3})$.

\textbf{Case 2:} both predictors are correct. Hence, $\LZeroOnearg = 0$ independently of the selector. It is immediate to check that $\LJAskarg \geq 0$ always, hence \[\frac{\LJAskarg}{\commonfrac}\geq 0 = \LJAskarg\]

\textbf{Case 3:} the selector $s(\mbx)=0$, $f(\mbx)\neq y$ and $g(\mbx,h)=y$.
Therefore, $\LZeroOnearg = 1$.
Now, consider that
\begin{multline}
    \LJAskarg =  
    \left(1-\ell^s\left(\tilde{s}(\mbx)\right)\right)\SL\left(\smfy,y\right)+\ell^s\left(\tilde{s}(\mbx)\right) \SL\left(\smgy, y\right) \geq \\
    \frac{1}{2}\SL\left(\smfy,y\right) + \frac{1}{2}\SL\left(\smgy, y\right) \geq
    \frac{1}{2}\SL\left(\nicefrac{1}{2}\right) + \frac{1}{2}\SL\left(\smgy, y\right) \geq \\
    \frac{1}{2}\SL\left(\nicefrac{2}{3}\right) + \frac{1}{2}\SL\left(\smgy, y\right),
\end{multline}
where the first inequality holds as $f(\mbx)\neq y$ and since $s(\mbx)=0$, therefore $\tilde{s}(\mbx)<0$ and $\ell^s\left(\tilde{s}(\mbx)\right)<\nicefrac{1}{2}$; and the second and third inequality holds as as $f(\mbx)\neq y$.
Now we can divide both sides of the inequality by $\commonfrac$, therefore we get that:
\begin{multline}
   \frac{\LJAskarg}{\commonfrac} \geq  \frac{\frac{1}{2}\SL\left(\nicefrac{2}{3}\right) + \frac{1}{2}\SL\left(\smgy, y\right)}{\commonfrac} \geq \\
   1+\frac{\frac{1}{2}\SL\left(\smgy, y\right)}{\commonfrac} \geq 1 = \LZeroOnearg
\end{multline}

\textbf{Case 4:} it is the symmetrical of Case 3, i.e., the selector $s(\mbx)=1$, $f(\mbx)= y$, and $g(\mbx)\neq y$.
Also in this case, $\LZeroOnearg = 1$.
Now, consider that
\begin{multline}
    \LJAskarg =  
    \left(1-\ell^s\left(\tilde{s}(\mbx)\right)\right)\SL\left(\smfy,y\right)+\ell^s\left(\tilde{s}(\mbx)\right) \SL\left(\smgy, y\right) \geq \\
    \frac{1}{2}\SL\left(\smfy, y\right) + \frac{1}{2}\SL\left(\smgy,y\right) \geq \frac{1}{2}\SL\left(\smfy, y\right) +
    \frac{1}{2}\SL\left(\nicefrac{1}{2}\right) \geq \\
     \frac{1}{2}\SL\left(\smfy, y\right) +\frac{1} {2}\SL\left(\nicefrac{2}{3}\right),
\end{multline}
where the first inequality holds as $g(\mbx,h)\neq y$ and since $s(\mbx)=1$, therefore $\tilde{s}(\mbx)>0$ and $\ell^s\left(\tilde{s}(\mbx)\right)>\nicefrac{1}{2}$; and the second and third inequality holds as as $g(\mbx,h)\neq y$.
Now we can divide both sides of the inequality by $\commonfrac$, therefore we get that:
\begin{multline}
   \frac{\LJAskarg}{\commonfrac} \geq  \frac{ \frac{1}{2}\SL\left(\smfy, y\right) + \frac{1}{2}\SL\left(\nicefrac{2}{3}\right)}{\commonfrac} \geq \\
   \frac{\frac{1}{2}\SL\left(\smfy, y\right)}{\commonfrac} +1 \geq 1 = \LZeroOnearg
\end{multline}

\textbf{Case 5:} the selector is $s(\mbx)=1$, $f(\mbx)\neq y$ and $g(\mbx,h)=y$.
It is immediate to see that in this case $\LZeroOnearg=0$.
Therefore, as shown before, $\LJAskarg\geq 0$ and the inequality holds when dividing both sides by $\commonfrac$.

\textbf{Case 6:} the selector is $s(\mbx)=0$, $f(\mbx)= y$ and $g(\mbx,h)\neq y$.
Also in this case, it is immediate to see that $\LZeroOnearg=0$.
Therefore, as shown before, $\LJAskarg\geq 0$ and the inequality holds when dividing both sides by $\commonfrac$.

The six cases show that $\frac{\tilde{L}^{ask}(\cdot)}{\commonfrac}\geq L^{ask}_{0-1}(\cdot)$ for any $\mbx,y,h\in\mcX\times\mcY\times\mcH$ and $f,g,s \in \mcF\times\mcG\times\mcS$.

Now consider a distribution and predictors such that there exists a zero error solution $f^*,g^*,s^* \in \mcF\times\mcG\times\mcS$ with $\mcE_{L^{ask}_{0-1}}=0$. Consider $\hat{f}, \hat{g},\hat{s}$ as the minimizer of the surrogate loss $\tilde{L}^{ask}$. i.e.,~$\hat{f}, \hat{g},\hat{s} \in \argmin_{f,g,s \in \mcF\times\mcG\times\mcS}\mcE_{\tilde{L}^{ask}}(f,g,s)$. Then the following inequality holds:

\begin{multline*}
    \mcE_{L^{ask}_{0-1}}(\hat{f}, \hat{g},\hat{s}) \leq \frac{\mcE_{\tilde{L}^{ask}}(\hat{f}, \hat{g},\hat{s})}{\commonfrac} \leq 
    \frac{\mcE_{\tilde{L}^{ask}}(\alpha f^*, \alpha' {g^*},s^*)}{\commonfrac} = \\
    \frac{1}{\commonfrac}\mathbb{E}\left[
    \tilde{L}^{ask}\left(\alpha \tf^*(\mbx), \alpha' \tg^*(\mbx,h), \gamma\tilde{s}^*(\mbx)\right) \mid s^*(\mbx) =0
    \right] P\left(s^*(\mbx) = 0\right) +\\
    \frac{1}{\commonfrac}\mathbb{E}\left[
    \tilde{L}^{ask}\left(\alpha \tf^*(\mbx), \alpha' \tg^*(\mbx,h), \gamma\tilde{s}^*(\mbx)\right) \mid s^*(\mbx) =1
    \right] P\left(s^*(\mbx) = 1\right) = \\
    \frac{1}{\commonfrac}\mathbb{E}[\SL\left(\sigma\left(\alpha \tf^*(\mbx)\right), y\right)\left(1-\ell^s\left(\gamma\tilde{s}^*(\mbx)\right)\right)
+\SL\left(\sigma\left(\alpha' \tg^*(\mbx)\right), y\right)\ell^s\left(\gamma\tilde{s}^*(\mbx)\right)
\mid s^*(\mbx) = 0] P(s^*(\mbx)=0) +\\
\frac{1}{\commonfrac}\mathbb{E}[\SL\left(\sigma\left(\alpha \tf^*(\mbx)\right), y\right)\left(1-\ell^s\left(\gamma\tilde{s}^*(\mbx)\right)\right)
+\SL\left(\sigma\left(\alpha' \tg^*(\mbx,h)\right), y\right)\ell^s\left(\gamma\tilde{s}^*(\mbx)\right)
\mid s^*(\mbx) = 1] P(s^*(\mbx)=1) \leq \\
    \frac{1}{\commonfrac}\mathbb{E}[\SL\left(\sigma\left(\alpha \tf^*(\mbx)\right), y\right)\left(1-\ell^s\left(\gamma\tilde{s}^*(\mbx)\right)\right)
+\ell^s\left(\gamma\tilde{s}^*(\mbx)\right)
\mid s^*(\mbx) = 0] P(s^*(\mbx)=0) +\\
\frac{1}{\commonfrac}\mathbb{E}[\left(1-\ell^s\left(\gamma\tilde{s}^*(\mbx)\right)\right)
+\SL\left(\sigma\left(\alpha' \tg^*(\mbx,h)\right), y\right)\ell^s\left(\gamma\tilde{s}^*(\mbx)\right)
\mid s^*(\mbx) = 1] P(s^*(\mbx)=1)
\end{multline*}

Where the first inequality holds by $\frac{\tilde{L}^{ask}(\cdot)}{\commonfrac}\geq L^{ask}_{0-1}(\cdot)$, the second inequality holds because $\hat{f}, \hat{g}, \hat{s}$ minimize $\mcE_{\tilde{L}^{ask}}(\cdot)$ and both $\mcF$ and $\mcG$ are closed under scaling, the two equalities hold by definition and the last inequality hold as the loss $\ell^{surr}$ is always lower or equal one by definition.

Now, conditional on $s^*(\mbx) = 0$,  we can split the limit into two parts, i.e.,
\[\lim_{\alpha,\alpha', \gamma\to \infty}\mathbb{E}[\SL\left(\sigma\left(\alpha \tilde{f}^*(\mbx)\right), y\right)\left(1-\ell^s\left(\gamma\tilde{s}^*(\mbx)\right)\right)\mid s^*(\mbx) = 0] P(s^*(\mbx)=0)\]
and 
\[\lim_{\alpha,\alpha', \gamma\to \infty}\mathbb{E}[\ell^s\left(\gamma\tilde{s}^*(\mbx)\right)\mid s^*(\mbx) = 0] P(s^*(\mbx)=0)\]
Now, by the monotone convergence theorem, the limits can be taken inside the expected value and computed component-wise. 
The first part goes to zero because 
$(i)$ $s^*(\mbx)=0$ implies $\tilde{s}^*(\mbx)<0$, thus $\lim_{\gamma\to\infty}1-\ell^s\left(\gamma\tilde{s}^*(\mbx)\right)=1$ by assumption; 
$(ii)$ $\lim_{\alpha\to\infty}\SL\left(\sigma\left(\alpha \tf^*(\mbx)\right), y\right)=0$ as $f^*$ is the optimal solution and $f$ is closed under scaling.
The second part goes to zero as $\lim_{\gamma\to\infty}\ell^s\left(\gamma\tilde{s}^*(\mbx)\right)=0$.

Similarly, the second term is such that conditional on $s^*(\mbx)=1$, we can reduce it to two parts, i.e., 
\[\lim_{\alpha,\alpha', \gamma\to \infty}\mathbb{E}[\left(1-\ell^s\left(\gamma\tilde{s}^*(\mbx)\right)\right)\mid s^*(\mbx) = 1] P(s^*(\mbx)=1)\]
and 
\[\lim_{\alpha,\alpha', \gamma\to \infty}\mathbb{E}[+\SL\left(\sigma\left(\alpha' \tg^*(\mbx,h)\right), y\right)\ell^s\left(\gamma\tilde{s}^*(\mbx)\right)\mid s^*(\mbx) = 1] P(s^*(\mbx)=1)\]
Now, by the monotone convergence theorem, the limits can be taken inside the expected value and computed component-wise. 
The first part goes to zero because 
$(i)$ $s^*(\mbx)=1$ implies $\tilde{s}^*(\mbx)>0$, thus $\lim_{\gamma\to\infty}1-\ell^s\left(\gamma\tilde{s}^*(\mbx)\right)=0$ by assumption; 
$(ii)$ $\lim_{\alpha\to\infty}\SL\left(\sigma\left(\alpha \tg^*(\mbx,h)\right), y\right)=0$.
The second part goes to zero as $\lim_{\gamma\to\infty}\ell^s\left(\gamma\tilde{s}^*(\mbx)\right)=0$ and $\lim_{\gamma\to\infty}\ell^s\left(\gamma\tilde{s}^*(\mbx)\right)=1$.

Combining the two, we have that $\mcE_{L^{ask}_{0-1}(\hat{f},\hat{g}, \hat{s})}=0$ hence $\tilde{L}^{ask}$ is realizable $\mcF\times\mcG\times\mcS$-consistent with respect to $L^{ask}_{0-1}$.

\section{Additional Experimental Details}
\label{app:further-experimental-details}

\paragraph{Training Details. }For \synth{}, we train all the methods using a fixed budget of 150 epochs. In greater detail, for \textsc{LtD} the model $f$ was trained for 150 epochs; for \Seq{}, we trained $g$ for 50 epochs and $f$ for the remaining 100 epochs; for \Joint{} we pretrained both $f$ and $g$ for 50 epochs independently using cross-entropy, and then train $f$, $g$ and $s$ jointly for the remaining 100 epochs using $\tilde{L}^{ask}$ as specified in the main paper. 

For \chx{} (both kinds of feedback), we train all the methods using a fixed budget of 300 epochs. More precisely, for \textsc{LtD} the model $f$ was trained for the whol 300 epochs; for \Seq{}, we trained $g$ for 200 epochs and $f$ for the remaining 100 epochs; for \Joint{} we pretrained both $f$ and $g$ for 200 epochs independently using cross-entropy, and then train $f$, $g$ and $s$ jointly for the remaining 100 epochs using $\tilde{L}^{ask}$ as specified in the main paper. 

For both datasets, we use Stochastic Gradient Descent as an optimizer, setting an initial learning rate of $.001$.
We observe that the independent pretraining phase for \Joint{} avoids training dynamics that collapse on either training only $f$ or $g$.

\paragraph{Hardware and Training Time.} For our experiments, we use a 224-core machine with Intel(R) Xeon(R) Platinum 8480CL CPU and eight NVIDIA H100 80GB HBM3, OS Ubuntu 22.04.3 LTS. 
For \synth{}, \textsc{LtD}'s training took on average $\approx 40  \pm .60$ seconds, \Seq{} $\approx 48 \pm .37$ seconds and \Joint{} $\approx 75 \pm 3$ seconds.
For \chx{} - \texttt{Unc Feedback}, \textsc{LtD}'s training took on average $\approx 4860  \pm 50$ seconds, \Seq{} $\approx 5010 \pm 289$ seconds and \Joint{} $\approx5956 \pm 130$ seconds.
For \chx{} - \texttt{LtD-Feedback}, \textsc{LtD}'s training took on average $\approx 4860\pm 50 $ seconds, \Seq{} $\approx 
5967\pm 298$ seconds and \Joint{} $\approx
6602 \pm 207$ seconds.

\end{document}